 \documentclass[pmlr,twocolumn,10pt]{jmlr} % W&CP article

\usepackage{booktabs}
\usepackage{authblk}
\usepackage[load-configurations=version-1]{siunitx} % newer version 

\usepackage{caption}
\usepackage{float}
\theorembodyfont{\upshape}
\theoremheaderfont{\scshape}
\theorempostheader{:}
\theoremsep{\newline}

\jmlrvolume{}
\jmlryear{2021}
\jmlrsubmitted{}
\jmlrpublished{}
\jmlrworkshop{Machine Learning for Health (ML4H) 2021} % W&CP title

\title[]{Self-Supervised Learning for 3D Medical Image Analysis using \titlebreak 3D SimCLR and Monte Carlo Dropout}

\author[1,*]{%
\centering
 Yamen Ali
}
\author[2, $\dagger$]{%
  Aiham Taleb
}
\author[1,*]{%
 Marina M.-C. Höhne
}
\author[2,3,$\dagger$]{%
  Christoph Lippert
}
\affil[*]{\texttt {\{firstname.lastname\}@tu-berlin.de}}
\affil[$\dagger$]{\texttt {\{firstname.lastname\}@hpi.de}}
\affil[1]{Technical University of Berlin, Germany}
\affil[2]{Hasso Plattner Institute for Digital Engineering, University of Potsdam, Germany}
\affil[3]{Hasso Plattner Institute for Digital Health at the Icahn School of Medicine at Mount Sinai, NYC, USA}

\begin{document}

\maketitle

\begin{abstract}
 Self-supervised learning methods can be used to learn meaningful representations from unlabeled data that can be transferred to supervised downstream tasks to reduce the need for labeled data. In this paper, we propose a 3D self-supervised method that is based on the contrastive (SimCLR) method. Additionally, we show that employing Bayesian neural networks (with Monte-Carlo Dropout) during the inference phase can further enhance the results on the downstream tasks. We showcase our models on two medical imaging segmentation tasks: i) Brain Tumor Segmentation from 3D MRI, ii) Pancreas Tumor Segmentation from 3D CT. Our experimental results demonstrate the benefits of our proposed methods in both downstream data-efficiency and performance.
\end{abstract}

\section{Introduction}
\label{sec:intro}

As 3D medical imaging became an essential tool in medicine, the need for accurate and reliable machine learning algorithms that analyze such images has become more apparent. However, acquiring sufficient amounts of annotated 3D images is a non-trivial task due to the challenges related to privacy issues or the sheer time and cost required to get expert annotations for such data. Hence, this motivates other solutions to address the scarcity of annotations.

Self-Supervised Learning (SSL) has proven to be a powerful technique that allows constructing meaningful representations for the images by applying pretext tasks on the unlabelled data, e.g. \cite{3d-methods,review_annotation_efficient,zhou2019models}. 
\citet{simclr} introduced  a novel \emph{Sim}ple framework for \emph{C}ontrastive \emph{L}earning of visual \emph{R}epresentations (SimCLR). SimCLR leverages the normalized temperature-scaled cross-entropy loss (NT-Xent) in order to maximize the similarity between latent space representations of various augmentations of the same data point. In their work, they show how the representations learned from SimCLR can achieve state-of-the-art results when used in the downstream tasks of 2D natural image classification.

\textbf{Our contributions:} We introduce a method to utilize SimCLR in volumetric 3D image segmentation. Our method addresses the inherent challenges in semantic segmentation on 3D scans such as imbalanced classes and the expensive process of data annotation. Moreover, we show that the additional inclusion of uncertainties through approximate Bayesian weight inference in the form of Monte Carlo (MC) Dropout in the self-supervised algorithm significantly enhances the performance on segmentation tasks. 

\section{Method}
Our approach consists of three main steps. We start by a self-supervised learning step that results in a trained encoder $g_{enc}$. In the second step, we fine-tune the encoder $g_{enc}$ with the downstream task, e.g. segmentation task, using the annotated data. Finally, in the third step, we apply MC Dropout during prediction and report the dice scores on the test data.

\subsection{Pretask Training}
In the following, we propose a pretask that generalizes SimCLR to volumetric 3D inputs such that the full 3D spatial context of the scans is explored. We start by randomly sampling a batch of $M$ 3D scans, then each 3D scan is split into $P$ equally-sized non-overlapping 3D patches resulting in $N = M*P$ input samples. Before processing the input by the model, two random composite augmentations (chosen from: 3D rotation with different angles on one or more axis, color distortion, identity, Gaussian noise, Gaussian blur, and Sobel filtering) are applied onto each 3D patch leading to a dataset size of $2N$. Hence, similar as in SimCLR there exists one positive pair for every input sample, i.e., one pair originating from the same original 3D patch, and $2(N-1)$ negative pairs, i.e., originating from different 3D patches.

%\subsubsection{Data Augmentation}
%Data augmentation plays a vital role in contrastive learning. Hence, we define the following augmentations on 3D patches: 3D rotation with different angles on one or more axis, color distortion, identity, Gaussian noise, Gaussian blur and Sobel filtering.

\subsubsection{Model Architecture \& Loss}
The model architecture used for the pretask is similar to the one proposed in \citet{simclr}, consisting of an  encoder $g_{enc}$ (3D-CNN) followed by a non-linear projection head (Dense layer). As loss function we used a normalized temperature-scaled cross entropy to compute the loss for a positive pair of two augmented 3D patches $z_i$ and $z_j$ in the latent space representations (i.e. dense layer output):
\begin{equation} \label{eq:nxt}
l_{i,j} = -log \frac{  \exp(sim(z_i,z_j)/\tau)}  {\sum_{k=1}^{2N} 1_{[k \neq i]}  \exp(sim(z_i,z_k)/\tau) }
\end{equation}
where $sim$ is the cosine similarity and $\tau$ the temperature parameter. %The overall loss is defined as:
%\begin{equation}
%L =  \frac{1} {2N} \sum_{k=1}^{N}  [ l_{2k-1,2k} + l_{2k,2k-1} ]
%\end{equation}

\subsection{Finetuning}
To perform the required downstream segmentation task we keep only the pretrained encoder $g_{enc}$ without the non-linear projection head. The encoder outputs are passed to a decoder $g_{dec}$ and the model, a U-Net proposed by \citet{unet}, is trained in a supervised manner using the annotated 3D scans. 
As a loss function, the weighted dice score is used in order to maximize the intersection-over-union ratio between the model predictions and the ground truth across all classes. Formally expressed as:
\begin{equation}\label{eq:dice}
Dice  = - \frac{1}{N} \sum_{i=1}^{N}  \frac{ 2(\sum_{j=1}^{M} Y_{ij} \times T_{ij}) + s}{ (\sum_{j=1}^{M} Y_{ij}  + \sum_{j=1}^{M}  T_{ij} ) + s}
\end{equation}

where $Y_{ij}$ is the probability that pixel $j$ belongs to class $i$, $T_{ij}$ is the ground truth indicator, which is 1 if pixel $j$ belongs to class $i$ and 0 otherwise, $s$ is a smoothing parameter (otherwise called epsilon), $N$ is the number of classes, and $M$ is the number of pixels.

\subsection{Bayesian Approximation} \label{sec:bayesianApproximation}
Usually the predicted segmentation mask is deterministic and we lack information about the models certainty of the pixel-wise predictions. However, as shown in \citet{monte}, MC dropout can be used as a Bayesian approximation method. Accordingly, we obtain uncertainty estimates on the predicted segmentation, by applying MC dropout during the testing phase. Descriptively, this can be seen as a procedure, where, instead of sampling from a learned posterior distribution on the weights, we apply dropout during testing, thus obtaining a subset of weights that define a sampled sub-network. Note that the number of networks we sample via dropout is a hyperparameter, that needs to be set. 
The predictions of the individual sub-networks follow the multimodality strategies learned in the network and can be aggregated in various ways to achieve more transparency regarding the models certainty. We investigate the following four different aggregation methods:
\begin{itemize}
\setlength\itemsep{-0.2em}
  \item \textbf{Majority:} A pixel labeled with the class, where the majority of the networks agreed on. 
  %is considered of $class_x$ if the majority of the networks predicted it as $class_x$
  \item \textbf{Weighted Majority:} Additionally to the Majority aggregation, each class is multiplied by a weight to enable a class specific prioritizing. 
  %Same as majority,but the votes are weighted (scaled) per class. E.g., in \figureref{fig:fig7} we weighted the votes as following ( $background = 1, Pancreas = 2, tumor = 2$ )
  \item \textbf{Borda:} 
  For each pixel-wise prediction, points are distributed, such that the class with the highest probability has the most points and the class with the lowest probability the fewest points. Then the points of all sub-networks are added pixel-wise and class label that has achieved the most points is assigned to the pixel. 
  %For each network, the class probabilities per pixel are sorted in descending order. The first class gets 2 points, the second 1 point, and so on. The predicted class has the most points.
  \item \textbf{Union per $class_x$:} Given a class of interest, a pixel is assigned to that class if at least one sub-network predicts this class for that pixel. 
  %the final prediction is considered to be $class_x$ if only one network predicts it for any pixel. 
  This prediction aggregation encourages the algorithm to be sensitive to specific classes, e.g. Tumor. However, such approach affects the overall dice score as it produces more false positives.
\end{itemize}

% which is summarized in algorithm~\ref{algo1}.
% \begin{algorithm}
% \floatconts
%   {alg:mcd}%
%   {\caption{Uncertainty Estimation via MC Dropout} \commentMH{I think we do not need an algorithm for explaining how we use MC Dropout - it takes a lot of space and I we might explain it better via text.}}%
% {%
% \begin{enumerate}
%   \item Enable dropout during prediction.
%   \item Predictions = []
%   \item For $k=1$ to maximum number of iterations
%         \begin{enumerate}
%             \item prediction = model.predict(data)
%             \item Predictions.append(prediction)
%         \end{enumerate}
%   \item The final annotation for each pixel is its most predicted class (i.e. majority voting)
% \end{enumerate}
% }% 
% \label{algo1}
% \end{algorithm}

%This algorithm simulates having multiple models (an ensemble), each iteration representing a model, with their weights drawn from the posterior distribution. 
%In the first step we can have different configurations, e.g. we can set different dropout values for the encoder and the decoder or different dropout values per dropout layer.
%Furthermore, a different voting protocol can be used at the final step, e.g. union over a class where a pixel is considered from $class_x$ if one or more model predict it to be from $class_x$. 

\section{Experimental Results \& Conclusion}
In this section, we show some initial results of applying our 3D SimCLR approach to two datasets: the Pancreas tumor segmentation dataset by \citet{pancrease} and the Brain tumor segmentation dataset by \citet{brats}. Then, we evaluate the impact of MC dropout on the model performance. 

\subsection{Pancreas Dataset Results}
This dataset from the medical Decathlon benchmark \citet{pancrease} contains 3D CT scans of Pancreas tumors. It consists of 420 scans with only 281 of them being annotated. Each scan consists of voxels from 3 classes: background, pancreas, and tumor.
All 420 scans are used during the pretask training phase. For the fine-tuning, we split the 281 annotated scans into a training set of 197 scans, and a test set of 84 scans, which is only used for evaluation.
\figureref{fig:fig1} shows the improvement obtained by using the pretask encoder during finetuning, when only up to 25\% of the labeled data is available. The Pancreas dataset is challenging in nature due to the fact that the Pancreas and tumor classes occupy only a small area of the scan, i.e. a class imbalance.

\begin{figure}
 % Caption and label go in the first argument and the figure contents
 % go in the second argument
\floatconts
  {fig:fig1}
  {\caption{Visualization of the average dice score (y-axis) for the pancreas test dataset when finetuning the model with 5\%, 10\%, 25\%, 50\%, 100\% of the training dataset (x-axis). Our proposed 3D SimCLR approach (blue line) outperforms the baseline approach (orange line) when less than 25\% of labeled data are available.
  %The blue line is a model where the encoder weights were obtained by our 3D SimCLR, and the orange is a model initialized randomly
  }}
  {\includegraphics[width=0.95\linewidth]{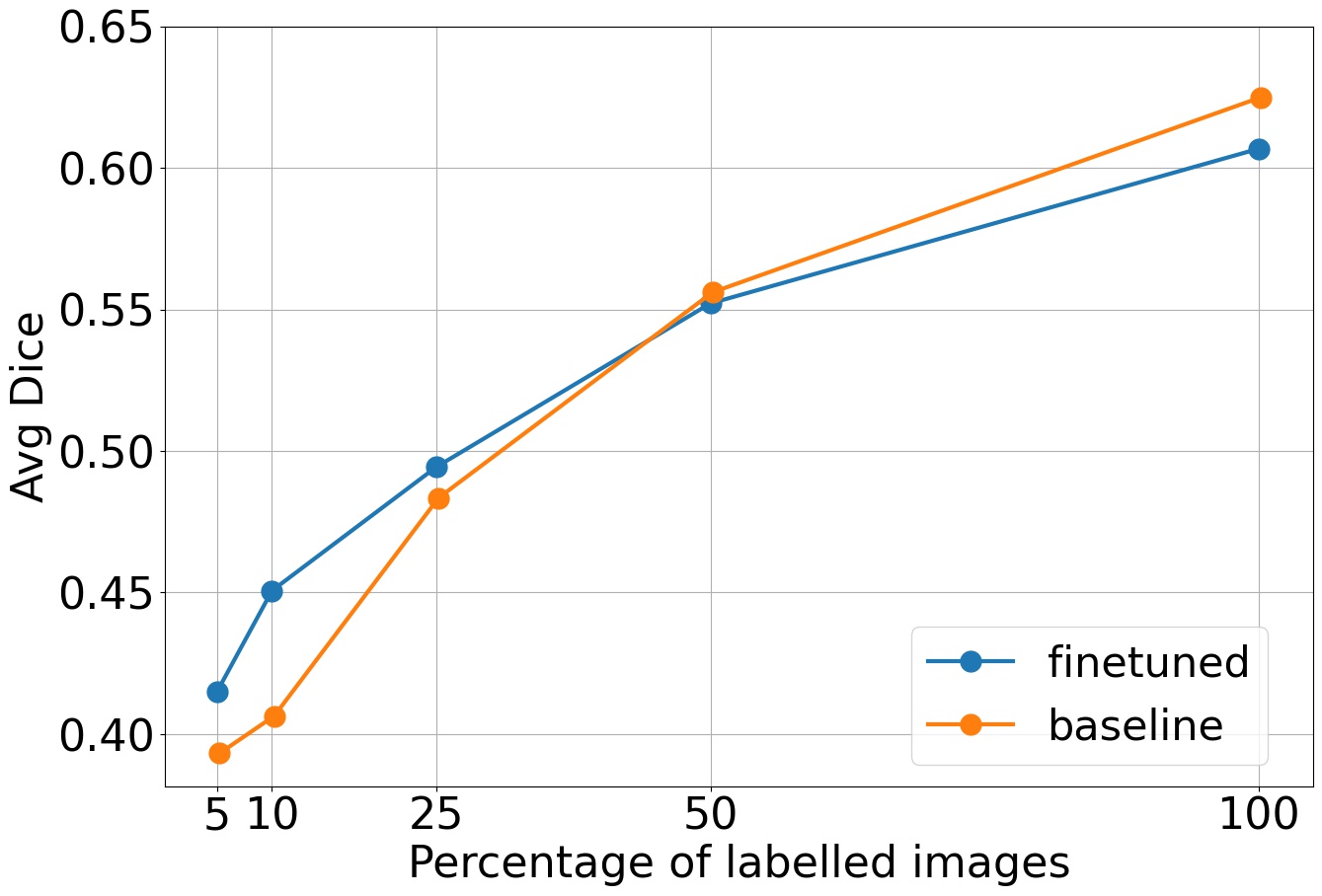}}
\end{figure}

\begin{figure}
 % Caption and label go in the first argument and the figure contents
 % go in the second argument
\floatconts
  {fig:fig3}
  {\caption{Average dice score of the Brats test dataset predictions when the model is fine-tuned using 5\%, 10\%, 25\%, 50\%, 100\% of the training dataset}}
  {\includegraphics[width=0.95\linewidth]{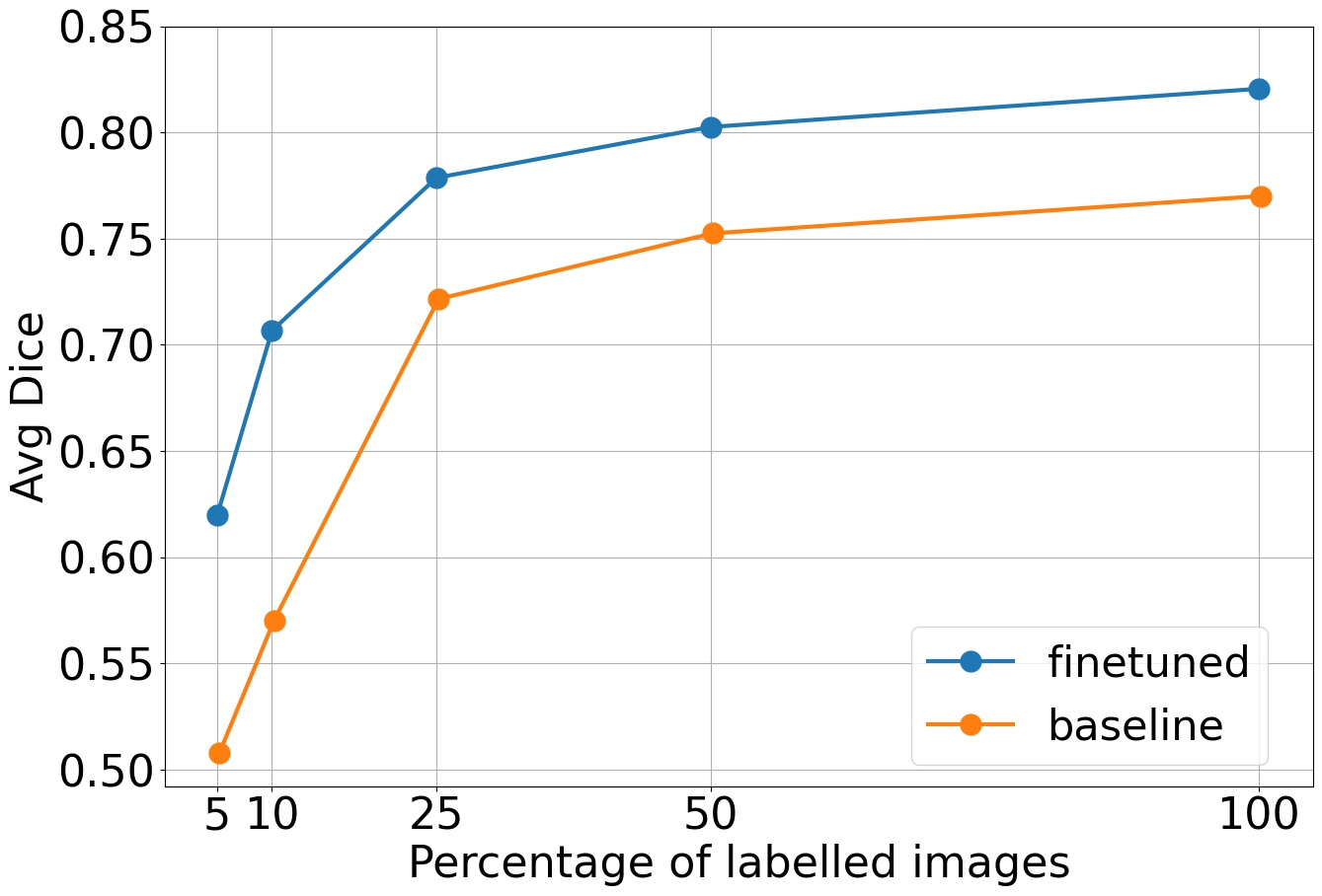}}
\end{figure}

% \begin{figure}
%  % Caption and label go in the first argument and the figure contents
%  % go in the second argument
% \floatconts
%   {fig:fig2}
%   {\caption{A 2D slice from a scan of the Pancreas dataset. The white area represents the Pancreas, the red area is the tumor, and the remaining pixels are background }}
%   {\includegraphics[width=0.5\linewidth]{images/Pancreas_example.png}}
% \end{figure}

\subsection{Brain Dataset Results}
The multimodal Brain Tumor Segmentation 2018 benchmark \citet{brats} dataset consists of 351 scans, out of which 285 are annotated. Each scan contains voxels from 4 classes: background, whole tumor, tumor core, and enhanced tumor. All 351 scans are used during the pretask training phase. For the fine tuning, we split the 285 annotated scans into a training set of 200 scans and a test set of 85 scans, which is only used for evaluation. The improvement obtained by using the pretask encoder in finetuning, as shown in \figureref{fig:fig3}, is more nuanced than in the Pancreas dataset. This can be attributed to the scans' nature where the tumor covers a larger area which reduces the effects of class imbalance.

% \begin{figure}
%  % Caption and label go in the first argument and the figure contents
%  % go in the second argument
% \floatconts
%   {fig:fig4}
%   {\caption{A 2D slice from a scan of the BraTS dataset. The green area represents the Whole Tumor, the red area is the tumor core, the blue area is the enhanced tumor, and the remaining pixels are background }}
%   {\includegraphics[width=0.5\linewidth]{images/brats_example.png}}
% \end{figure}

\subsection{Monte-Carlo Dropout Results}
\figureref{fig:fig4} shows that applying MC dropout improves the accuracy of the fine-tuned model, while the baseline model accuracy hardly changes. We conjecture this to be because the weights of the fine-tuned encoder, sampled by MC dropout, contain more useful features than their counterparts in the baseline's encoder. To verify the previous hypothesis, we run the MC dropout algorithm on the fine-tuned model with dropout enabled only on the encoder, only on the decoder, and on both of them. 
As \figureref{fig:fig5} shows, the major improvement of the dice score occurs only when dropout is performed on the encoder. The decoder alone is not sufficient to obtain such an improvement. \figureref{fig:fig6} shows how different dropout rates affect the improvement due to the MC Dropout algorithm. It also shows that a higher dropout rate, e.g. 0.5, adversely affects the results as it would introduce more noise during the inference.

\begin{figure}
 % Caption and label go in the first argument and the figure contents
 % go in the second argument
\floatconts
  {fig:fig4}
  {\caption{Comparison between the Pancreas fine-tuned and baseline models average dice scores before and after applying MC Dropout with 100 Iterations, encoder dropout=0.3, decoder dropout=0.0, and the majority voting protocol  }}
  {\includegraphics[width=0.95\linewidth]{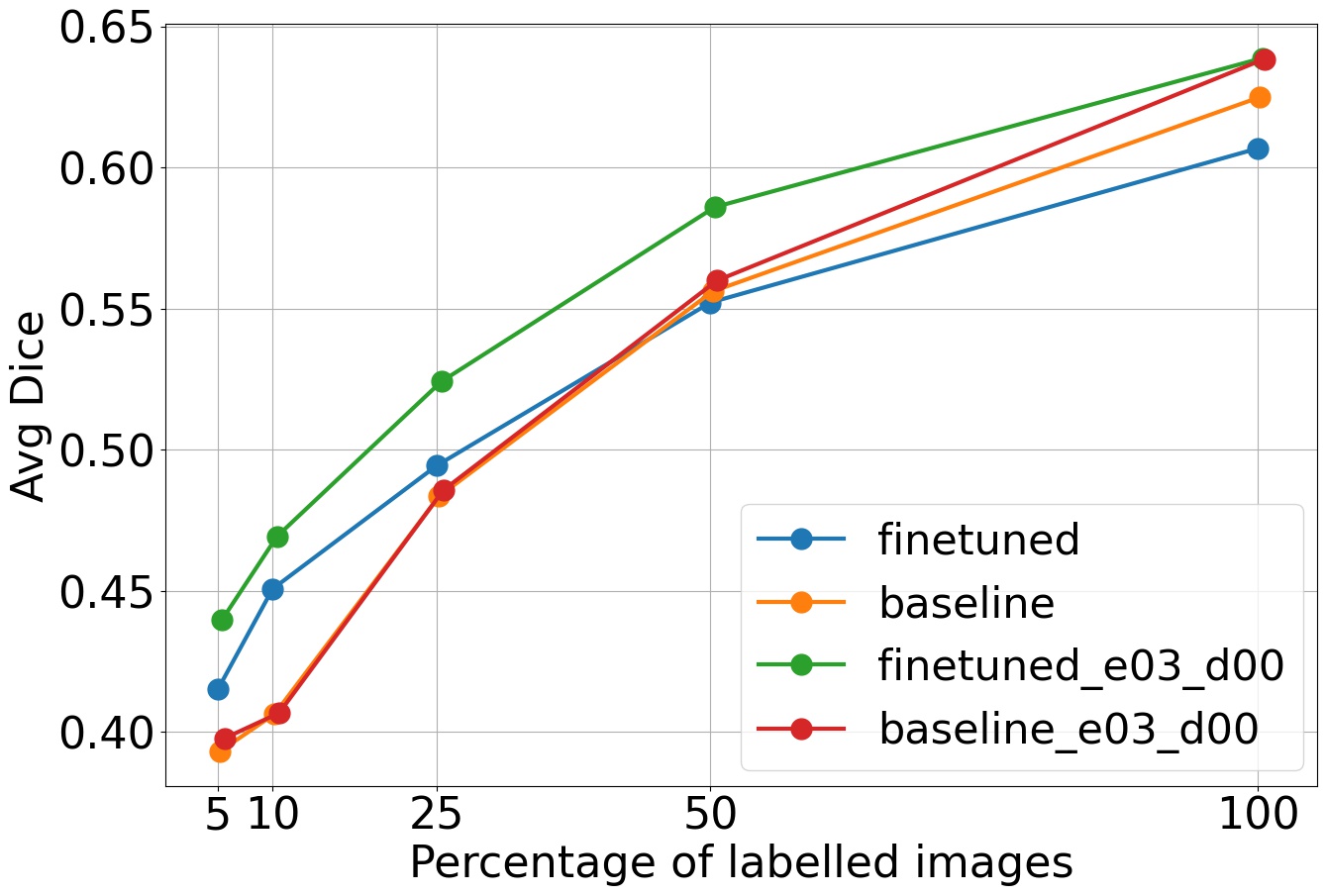}}
\end{figure}

\begin{figure}
 % Caption and label go in the first argument and the figure contents
 % go in the second argument
\floatconts
  {fig:fig5}
  {\caption{Comparison of MC dropout configurations in terms of average dice score when fine-tuning on Pancreas}}
  {\includegraphics[width=0.95\linewidth]{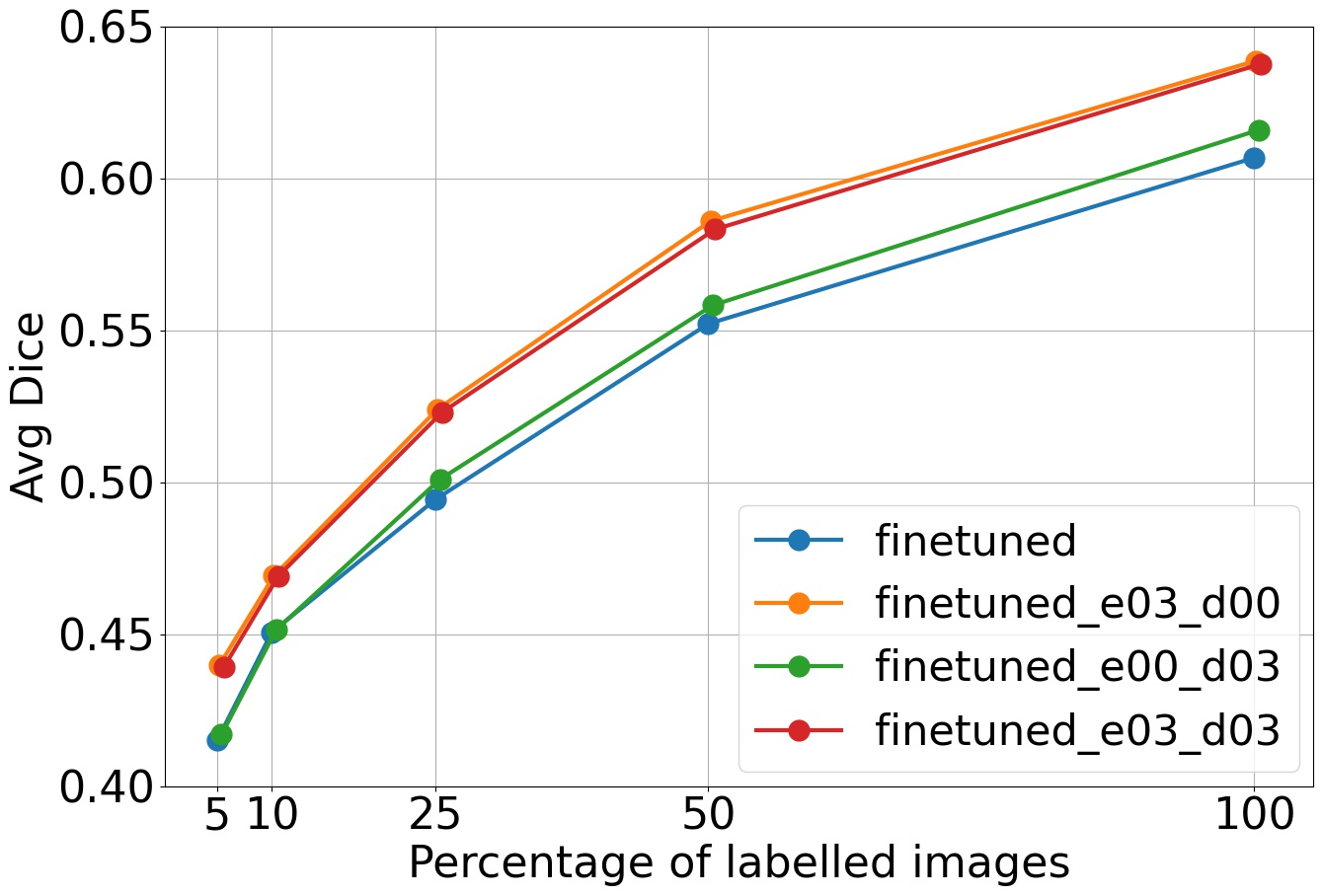}}
\end{figure}

\begin{figure}
 % Caption and label go in the first argument and the figure contents
 % go in the second argument
\floatconts
  {fig:fig6}
  {\caption{Comparison of different dropout rates in terms of average dice score when fine-tuning models on Pancreas }}
  {\includegraphics[width=0.95\linewidth]{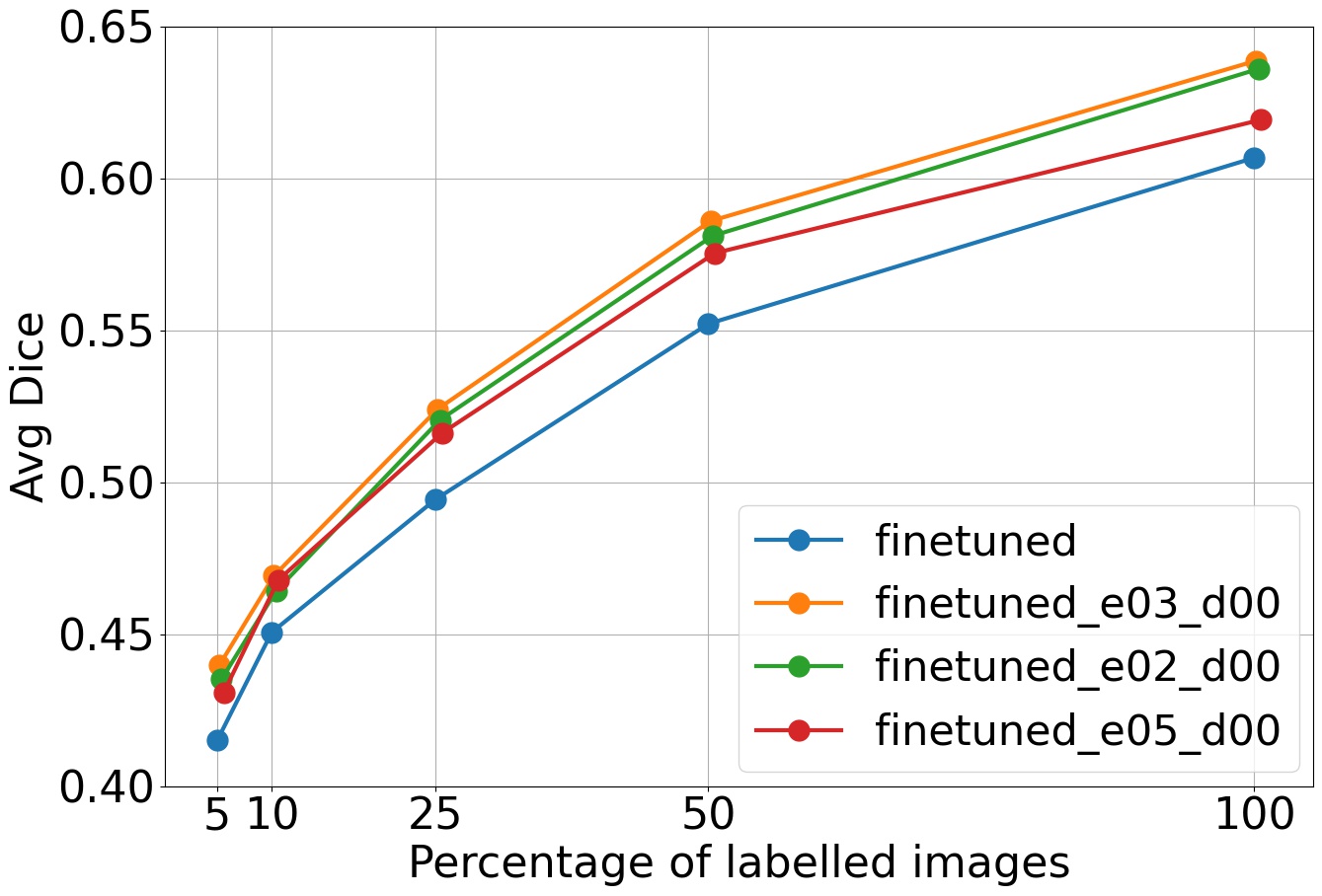}}
\end{figure}

% \begin{figure}
%  % Caption and label go in the first argument and the figure contents
%  % go in the second argument
% \floatconts
%   {fig:fig6}
%   {\caption{Comparison of different dropout rates in terms of average dice score when fine-tuning models on Pancreas }}
%   {\includegraphics[width=0.95\linewidth]{images/finetuned_dropout.jpg}}
% \end{figure}

In \figureref{fig:fig7}, we compare the different prediction aggregations presented in Section \ref{sec:bayesianApproximation} on the Pancreas dataset with encoder dropout rate $0.3$ and decoder dropout $0.0$. We can observe that these aggregation methods have a similar performance, whereby the Weighted Majority ($background = 1, Pancreas = 2, tumor = 2$ ) performs slightly better when over 5\% of labeled data are available.
The Borda and Majority protocols give almost identical results, indicating that the network predictions strongly agree.

\begin{figure}
 % Caption and label go in the first argument and the figure contents
 % go in the second argument
\floatconts
  {fig:fig7}
  {\caption{Comparison of the different prediction aggregation methods presented in Section \ref{sec:bayesianApproximation}. 
  %voting protocols in terms of avg. dice score when fine-tuning on Pancreas with encoder dropout 0.3 and decoder dropout 0. Borda and Majority protocols give almost identical results, indicating that the network predictions highly agree.
  }}
  {\includegraphics[width=0.95\linewidth]{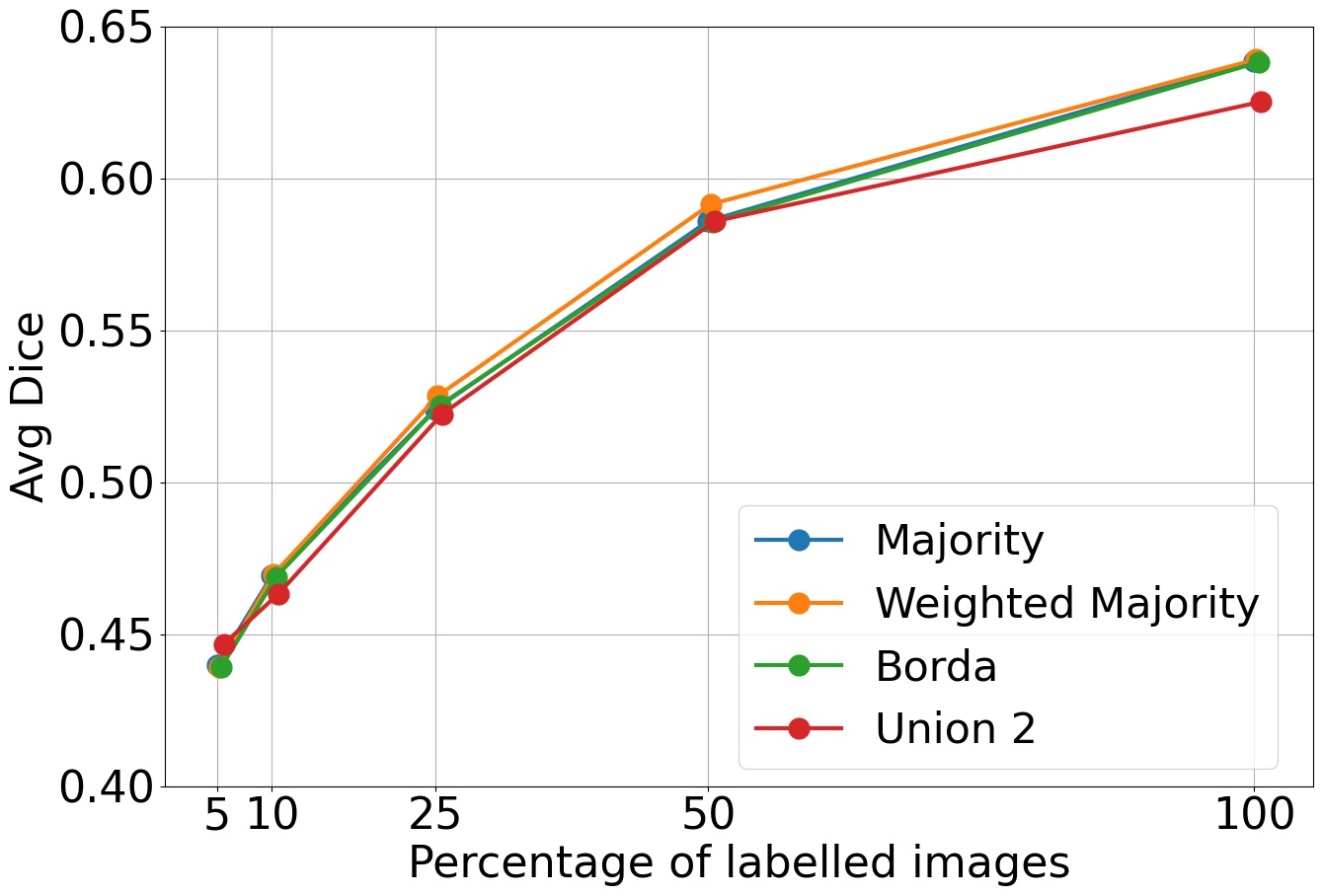}}
\end{figure}

% As mentioned before, different voting protocols can be used in the MCD algorithm. In \figureref{fig:fig7}, we compare the following ones:

% \begin{itemize}
% \setlength\itemsep{-0.2em}
%   \item \textbf{Majority:} A pixel is considered of $class_x$ if the majority of the networks predicted it as $class_x$
  
%   \item \textbf{Weighted Majority:} Same as majority, but the votes are weighted (scaled) per class. E.g., in \figureref{fig:fig7} we weighted the votes as following ( $background = 1, Pancreas = 2, tumor = 2$ )
  
%   \item \textbf{Borda:} For each network, the class probabilities per pixel are sorted in descending order. The first class gets 2 points, the second 1 point, and so on. The predicted class has the most points.
  
%   \item \textbf{Union per $class_x$:} The final prediction is considered to be $class_x$ if only one network predicts it for any pixel. This protocol encourages the algorithm to be sensitive to specific classes, e.g. Tumor. However, such approach affects the overall dice score as it produces more false positives.
% \end{itemize}
%
\begin{figure}[h]
 % Caption and label go in the first argument and the figure contents
 % go in the second argument
\floatconts
  {fig:fig8}
  {\caption{Heat-maps of different percentiles of the WT class predictions for a sample from the BraTS dataset. The black pixels represent the true WT.}}
  {\includegraphics[width=0.99\linewidth]{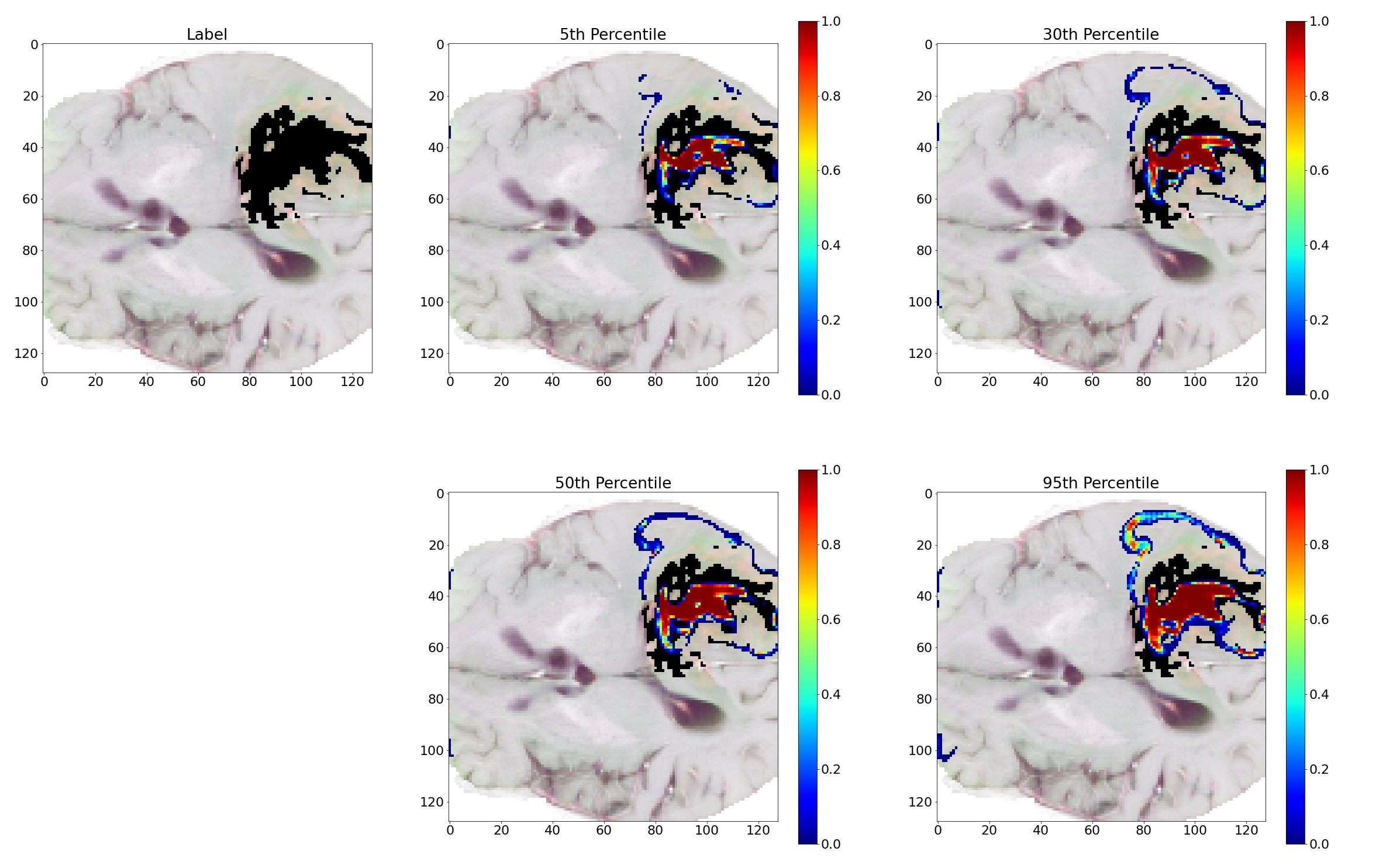}}
\end{figure}
Finally, in order to visualize the uncertainty knowledge gained by using MC dropout as a Bayesian approximation method, we compute pixel-wise different percentiles of the predictions and plot them as a heatmap over the Whole Tumor(WT) segmentation (black area) as shown in \figureref{fig:fig8} for the BraTS scan, with 100 MC Dropout steps (100 networks) similar as in \cite{bykov2021explaining}.
%We can observe that the sub-networks agree the most on the middle part of the WT as can be seen by the 5th percentile plot.
%effect of MC Dropout, we compare the heat-maps resulting from taking different percentiles of a specific class prediction, as in \figureref{fig:fig8}. 
%We can extract the most certain area of WT (red) from the 5th percentile, where 95\% of the networks agree on.  
Those heatmaps can be interpreted as a confidence indication that MC dropout networks have on pixel assignments, where the 5th percentile heatmap indicates the most certain area for WT (red=certain, blue=uncertain) as being in the middle of the WT area. From the 95th percentile heat map, we can observe a higher overlap with the true WT segmentation, however, the false positive rate increased as expected.

In conclusion, our experimental results demonstrate the potential of our proposed 3D SimCLR method, nested with additional models uncertainty information gained by Bayesian approximation at inference time, in terms of downstream data efficiency and performance improvement. These findings encourage further research in this direction, especially for annotation-starved healthcare applications.

%in case of the 5th percentile at least 95\% of the networks agree on the 

%\figureref{fig:fig8} shows the Whole Tumor (WT) class of a BraTS scan, with 100 MC Dropout steps (100 networks).
%In the $5^{th}$ percentile, we plot for each pixel the lowest probability of it being from class WT that 95\% of the networks agree on i.e. 95\% of the networks assign a probability for a pixel being a whole tumor that is higher or equal to the plotted value. Hence, the pixels with high heat values (in red) are the ones with high probability for class WT across 95\% of the networks.

% \acks{Acknowledgements go here.}

\bibliography{jmlr-sample}

\clearpage
\appendix 
\section{Datasets Samples} \label{apd:first}

\begin{figure}[H]
\centering
\subfigure{ %{.5\linewidth}
  \centering
  \includegraphics[width=0.4\linewidth]{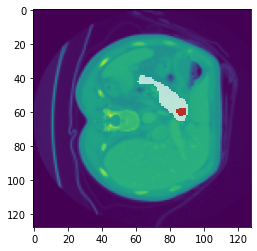}
  \label{fig:fig2a}
} \qquad
\subfigure{%{.5\linewidth}
  \centering
  \includegraphics[width=0.4\linewidth]{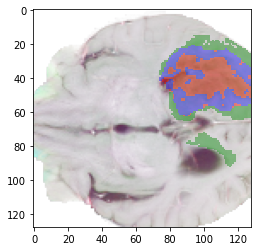}
  \label{fig:fig2b}
}
\caption{(a) A 2D slice from a Pancreas scan. The white area is the Pancreas, red is the tumor, and the rest is background. (b) A 2D slice from a BraTS scan. The green area is the whole tumor, red is the tumor core, blue is the enhanced tumor, and the rest is background
}
\end{figure}

\figureref{fig:fig2a} and \figureref{fig:fig2b} show a slice in Pancreas and BraTS 3D scans. We notice how the Pancreas tumor region is smaller than the Brain tumor region which makes the segmentation task on the Pancreas dataset intrinsically more challenging.  

\section{Training Details}  \label{apd:second}
\subsection{Data preprocessing}

As a preprocessing step on both datasets, a bounding box which surrounds the organs along each axis is found and all voxels outside this box are cropped out in order to reduce the amount of background voxels. Furthermore, all scans are resized into a unified resolution of $128 \times 128 \times 128$.

\subsection{Pretask Training}
For the temperature value in the \equationref{eq:nxt} we experimented with the same values suggested by \citet{simclr}, namely (0.05, 0.1, 0.5). 
For all the previously mentioned self supervised trained encoders, the temperature was set to (0.05) and the training ran for 1000 epochs. 

\subsection{Fine-tuning}
Both the fine-tuned model and the baseline model were trained for 400 epochs. But for the fine-tuned model, we follow a warm-up procedure suggested by \cite{3d-methods} where we freeze the encoder weights for the first 25 epochs. The smoothing value in \equationref{eq:dice} is set to \num[scientific-notation=true]{0.00001}.

% \begin{figure}
%  % Caption and label go in the first argument and the figure contents
%  % go in the second argument
% \floatconts
%   {fig:fig6}
%   {\caption{Comparison of different dropout rates in terms of average dice score when fine-tuning models on Pancreas }}
%   {\includegraphics[width=0.95\linewidth]{images/finetuned_dropout.jpg}}
% \end{figure}

\end{document}